\documentclass[twoside]{article}
\usepackage{aistats2e}
\usepackage{amsmath}
\usepackage{natbib}
\begin{document}

%

%

\twocolumn[

\aistatstitle{Joint Training of Deep Boltzmann Machines for Classification}

\aistatsauthor{Ian J. Goodfellow \And Aaron Courville \And Yoshua Bengio}

\aistatsaddress{ Universit\'e de Montr\'eal } ]

\begin{abstract}
We introduce a new method for training deep Boltzmann machines jointly.
Prior methods require an initial learning pass that trains
the deep Boltzmann machine greedily, one layer at a time, or do not
perform well on classification tasks.
\end{abstract}

\section{Deep Boltzmann machines}

A deep Boltzmann machine \citep{Salakhutdinov2009} is a probabilistic model
consisting of many layers of random variables, most of
which are latent. Typically, a DBM contains a set of $D$ input features $v$ that are called the {\em visible units}
because they are always observed during both training and evaluation.
The DBM is usually applied to classification problems and thus often represents the class label with a one-of-$k$ code
in the form of a discrete-valued label unit $y$. $y$  is observed (on examples for which it is
available) during training.
The DBM also contains several hidden units, which are usually organized into $L$ layers
$h^{(i)}$ of size $N_i, i=1,\dots,L$,with each unit in a layer conditionally independent of the other units in the layer given
the neighboring layers. These conditional independence properties allow fast Gibbs sampling because an entire layer of
units can be sampled at a time. Likewise, mean field inference with fixed point equations is fast because each fixed
point equation gives a solution to an entire layer of variational parameters.

A DBM defines a probability distribution by exponentiating and normalizing an energy function
\[
P(v,h,y) = \frac{1}{Z} \exp\left( -E(v,h,y) \right) \]
where
\[
Z = \sum_{v',h',y'} \exp \left( -E(v', h', y') \right). \]

$Z$, the partition function, is intractable, due to the summation over all possible states. Maximum likelihood
learning requires computing the gradient of $\log Z$. Fortunately, the gradient can be estimated using an MCMC
procedure \citep{Younes1999,Tieleman08}. Block Gibbs sampling of the layers makes this procedure efficient.

The structure of the interactions in $h$ determines whether further approximations are necessary. In the pathological
case where every element of $h$ is conditionally independent of the others given the visible units, the DBM is simply
an RBM and $log Z$ is the only intractable term of the log likelihood.
In the general case, interactions between different elements of $h$ render the posterior $P(h \mid v, y)$ intractable.
\citet{Salakhutdinov2009} overcome this
by maximizing the lower bound on the log likelihood given by the mean field approximation to
the posterior rather than maximizing the log likelihood itself. Again, block mean field inference over the layers makes
this procedure efficient.

An interesting property of the DBM is that the training procedure thus involves {\em feedback connections} between the
layers. Consider the simple DBM consisting of all binary valued units, with the energy function
\[ E(v,h) = - v^T W^{(1)} h^{(1)}  -h^{(1)T} W^{(2)} h^{(2)}. \]
Approximate inference in this model involves repeatedly applying two fixed-point update equations to solve for the
mean field approximation to the posterior. Essentially it involves running a recurrent net in order to obtain approximate
expectations of the latent variables.

Beyond their theoretical appeal as a deep model that admits simultaneous training of all components using a generative
cost, DBMs have achieved excellent performance in practice. When they were first introduced, DBMs set the state of the
art on the permutation-invariant version of the MNIST handwritten digit recognition task at 0.95.
(By permutation-invariant, we
mean that permuting all of the input pixels prior to learning the network should not cause a change in performance, so
using synthetic image distortions or convolution to engineer knowledge about the structure of the images into the system
is not allowed). Recently, new techniques were used in conjunction with DBM pretraining to set a new state
of the art of 0.79 \% test error \citep{DBLP:journals/corr/abs-1207-0580}.

\section{The joint training problem}

Unfortunately, it is not possible to train a deep Boltzmann machine using only the varational
bound and approximate gradient described above. \citet{Salakhutdinov2009} found that instead it must
be trained one layer at a time, where each layer is trained as an RBM. The RBMs can then be
modified slightly, assembled into a DBM, and the DBM may be trained with the learning rule
described above.

In this paper, we propose a method that enables the deep Boltzmann machine to be jointly trained.

\subsection{Motivation}

As a greedy optimization procedure, layerwise training may be suboptimal.
Recent small-scale experimental work has demonstrated this to be the case for
deep belief networks \citep{2012arXiv1212.1524A}.

In general, for layerwise training to be optimal, the training procedure for
each layer must take into account the influence that the deeper layers will
provide. The standard training procedure simply does not attempt to be optimal,
while the procedure advocated by \citep{2012arXiv1212.1524A} makes an optimistic
assumption that the deeper layers will be able to implement the best possible
prior on the current layer's hidden units. This approach does not work for
deep Boltzmann machines because the interactions between deep and shallow units
are symmetrical. Moreover, model architectures incorporating design features
such as sparse connections, pooling, or factored multilinear interactions make
it difficult to predict how best to structure one layer's hidden units in order
for the next layer to make good use of them.

\citet{Montavon2012arxiv} showed that reparameterizing the DBM to improve the condition
number of the Hessian results in succesful generative training without a greedy
layerwise pretraining step. However, this method has never been shown to have
good classification performance, possibly because the reparameterization makes
the features never be zero from the point of view of the final classifier.

\subsection{Obstacles}

Many obstacles make DBM training difficult. As shown by \citet{Montavon2012arxiv},
the condition number of the Hessian is poor
when the model is parameterized as having binary states.

Many other obstacles exist. The intractable objective
function and the great expense of methods of approximating it such as AIS makes it
too costly do line searches or early stopping. The standard means of approximating
the gradient are based on stateful MCMC sampling, so any optimization method that
takes large steps makes the Markov chain and thus the subsequent gradient estimates
invalid.

\section{The JDBM criterion}

Our basic approach is to use a deterministic criterion so that each of the above
obstacles ceases to be a problem.

Our specific deterministic criterion we call the {\em Joint DBM inpainting} criterion,
given by

\[ J(v, \theta) = \sum_i \log Q^*_i ( v_{S_i} ) \]

where
\[ Q^*(S_i) = \text{argmin}_Q D_{KL} \left( Q(v_{S_i} ) \Vert P( h \mid v_{-S_i} ) \right) . \]

This can be viewed as a mean field approximation to the generalized pseudolikelihood.
We backprop through the minimization of $Q$, so this can be viewed as training a family of
recurrent nets that all share parameters but each optimize a different task.

While both pseudolikelihood and likelihood are asymptotically consistent estimators, their
behavior in the limited data case is different. Maximum likelihood should be better for drawing
samples, but generalized pseudolikelihood can often be better for training a model to answer
queries conditioning on sets similar to the $S_i$ used during training. We view our work as similar
to \citep{Stoyanov2011}. The idea is to train the DBM to be a general question answering machine, using
the same approximations at train time as will be required at test time, rather than to train it to
be a good at generating MCMC samples that resemble the training data.

We train using nonlinear conjugate gradient descent on large minibatches of data. For each data point,
in the minibatch, we sample only one subset $S_i$ to train on, rather than attempting to sum over all
subsets $S_i$. We choose each variable in the model to be conditioned on independently from the others
with probability $p$. High values of $p$ work best, since the mean field assumption is applied to the
variables that are not selected to be conditioned on, and the more of those there are the worse the
mean field assumption is.

\subsection{MNIST experiments}

We used the MNIST dataset as a benchmark to compare our training method to the layerwise method
proposed by \citet{Salakhutdinov2009}. In order to replicate their technique as closely as possible we
refer to the accompanying demo code (\texttt{http://www.mit.edu/~rsalakhu/DBM.html}) rather than the paper itself. Since many important
details of the code are not included in the paper, we provide a summary of the code here.

\subsubsection{Prior method}
The demo code trains a DBM consisting of $v$, $h^{(1)}$, $h^{(2)}$, and $y$. This is accomplished in
three steps: 1) Training an RBM consisting of $v$ and $h^{(1)}$ to maximize the likelihood of $v$.
2) Training an RBM consisting of $h^{(1)}$, $h^{(2)}$, and $y$ to maximize the likelihood of $y$ and
$h^{(1)}$ when $h^{(1)}$ is drawn from the first RBM's posterior. 3) Assembling the RBMs into a DBM
and training it to maximize the variational lower bound on $\log P(v,y)$.

Thus far the model has only been trained generatively, though the labels $y$ are included. Its discriminative
performance--its ability to predict $y$ from $v$ is thus somewhat limited. We used mean field inference
to approximate $P(y \mid v)$ in the trained model and obtained a test set error of 2.15 \%.

In order to obtain better discriminative performance, the DBM is used to define a feature extractor / classifier
pipeline.

First, the dataset is augmented with features $\phi$. $\phi$ is computed once at the start of discriminative
training and then fixed, i.e., the discriminative learning does not change the value of $\phi$.
$\phi(v)$ is defined to be the mean field parameter vector  $\hat{h}^{(2)}$ obtained by
running mean field on $v$ with $\hat{y}$ clamped to 0. No explanation is given for clamping $\hat{y}$ to 0 in the code or the paper,
but we observe that it greatly improves generalization performance, even though it does not correspond to a standard
probabilistic operation like marginalizing out $y$.

Next, these features are fed into a multilayer perceptron that 
resembles one more step of inference:
\begin{align*}
\hat{h}^{(1)'} = \sigma \left( v^T A + f^T B + b^{(1)} \right) \\
\hat{h}^{(2)'} = \sigma \left( \hat{h}^{(1)'T} C + b^{(2)} \right) \\
\hat{y} = \text{softmax}\left( \hat{h}^{(2)'T} D \right)
\end{align*}

$A$, $B$, $C$, and $D$ are initialized to $W^{(1)}$, $W^{(2)T}$, $W^{(2)}$, and $W^{(3)}$,
respectively. They are then treated as independent parameters, i.e., $C$ is not constrained
to remain equal to the transpose of $D$ during learning. The MLP is finally trained to
maximize the log probability of $y$ under $\hat{y}$ using 100 epochs of nonlinear conjugate
gradient descent.

\subsection{Our method}

We follow the pre-existing procedure as closely as possible. The differences are as follows:
\begin{enumerate}
\item We do not have a layerwise pretraining phase.
\item When training the DBM over $v$, $h^{(1)}$, $h^{(2)}$ and $y$, we use the JDBM inpainting criterion
instead of PCD.
\item Rather than running training for a hard-coded number of epochs as in the DBM demo, we use
early stopping based on the validation set error. We use the first 50,000 training examples for
training and the last 10,000 for validation. After the validation set error starts to increase,
we train on the entire MNIST training set until the log likelihood on the last 10,000 examples
matches the log likelihood on the first 50,000 at the time that the validation set error began
to rise.
\end{enumerate}

We obtain a test set accuracy of 1.19 \% on MNIST.

We observe that a DBM trained with layerwise RBM pretraining followed by standard DBM variational
learning obtains a lower inpainting error on the training set than our models jointly trained using
the inpainting criterion. This suggests that our criterion correctly ranks models according to their
value as a classifier, but that our optimization procedure needs to be improved.

For comparison, our best result using standard DBM variational learning but without layerwise pretraining
was 1.69 \% test error. Using the centering trick, this increased to 2.03 \%. Both of these numbers are
likely to improve somewhat with more hyperparameter exploration.

\bibliography{strings,strings-shorter,ml,aigaion-shorter}
\bibliographystyle{natbib}

\end{document}